\def\year{2020}\relax
\documentclass[letterpaper]{article} 
\usepackage{aaai20}  
\usepackage{times}  
\usepackage{helvet} 
\usepackage{courier}  
\usepackage[hyphens]{url}  
\usepackage{graphicx} 
\usepackage{graphicx}  
\nocopyright

\usepackage[utf8]{inputenc}
\usepackage{algorithm}
\usepackage[noend]{algpseudocode}
\usepackage{mathtools}
\usepackage{amsmath}
\usepackage{booktabs}
\usepackage{tabularx}
\usepackage[dvipsnames]{xcolor}
\usepackage{subcaption}

\newcommand{\methodname}{DeepSmartFuzzer}

\title{\methodname: Reward Guided Test Generation For Deep Learning}

\author{Samet Demir\textsuperscript{\rm 1}, Hasan Ferit Eniser\textsuperscript{\rm 2}\thanks{Most of the work was done when the author was in Boğaziçi University.}, Alper Sen\textsuperscript{\rm 1} \\ 
\textsuperscript{\rm 1}Department of Computer Engineering, Boğaziçi University, Turkey\\
\textsuperscript{\rm 2}Max Planck Institute for Software Systems, Germany\\
samet.demir1@boun.edu.tr, hfeniser@mpi-sws.org, alper.sen@boun.edu.tr 
}

\algblock{Input}{EndInput}
\algnotext{EndInput}
\algblock{Output}{EndOutput}
\algnotext{EndOutput}

\newcommand{\citet}[1]{\citeauthor{#1} \shortcite{#1}}

\begin{document}

\maketitle

\begin{abstract}
Testing Deep Neural Network (DNN) models has become more important than ever with the increasing usage of DNN models in safety-critical domains such as autonomous cars. The traditional approach of testing DNNs is to create a test set, which is a random subset of the dataset about the problem of interest. This kind of approach is not enough for testing most of the real-world scenarios since these traditional test sets do not include corner cases, while a corner case input is generally considered to introduce erroneous behaviors. Recent works on adversarial input generation, data augmentation, and coverage-guided fuzzing (CGF) have provided new ways to extend traditional test sets. Among those, CGF aims to produce new test inputs by fuzzing existing ones to achieve high coverage on a test adequacy criterion (i.e. coverage criterion). Given that the subject test adequacy criterion is a well-established one, CGF can potentially find error inducing inputs for different underlying reasons. In this paper, we propose a novel CGF solution for structural testing of DNNs. The proposed fuzzer employs Monte Carlo Tree Search to drive the coverage-guided search in the pursuit of achieving high coverage. Our evaluation shows that the inputs generated by our method result in higher coverage than the inputs produced by the previously introduced coverage-guided fuzzing techniques.
\end{abstract}

\section{Introduction}
Given enough amount of data and processing power, training a Deep Neural Network (DNN), also called Deep Learning, is the most popular way for dealing with many hard computational problems such as image classification~\cite{CiresanMS2012}, natural language processing~\cite{SutskeverVQ2014} and speech recognition~\cite{HintonDYD2012}. Impressive achievements in such tasks raised expectations for deploying DNNs in real-world applications, including the ones in safety-critical domains. Examples of safety-critical applications include air traffic control~\cite{julian2016policy}, medical diagnostics~\cite{litjens2017survey} and autonomous vehicles~\cite{bojarski2016end}.

Despite the remarkable achievements, recent works \cite{szegedy2013intriguing,goodfellow2015explaining} have demonstrated that the DNNs are vulnerable to small perturbations on seed inputs, also called adversarial attacks. Considering the catastrophic results that can emerge from erroneous behavior in safety-critical systems, DNNs must be characterized by a high degree of dependability before being deployed in safety-critical systems. Furthermore, it is also suggested that DNN-based systems should be allowed for use in the public domain only after presenting high levels of trustworthiness \cite{BurtonGH2017}.

Testing is the primary practice for analyzing and evaluating the quality of a software system~ \cite{jorgensen2013software}. It helps in reducing the risk by finding and eliminating erroneous behaviors before deployment of the systems. Moreover, it provides evidence for the required levels of safety of the subject system. One of the most fundamental testing concepts is defining a coverage criterion, also called a test adequacy criterion, for a given test set. A coverage criterion divides the input space into equivalence classes and is satisfied if there exists at least one input for each equivalence class in the given test set. Having a test set that is satisfying a coverage criterion provides a degree of dependability to the system under test.

Recent research in the DNN testing area introduces new DNN-specific coverage criteria such as statement coverage inspired neuron coverage~\cite{pei2017deepxplore} and its variants~\cite{MJX18},
MC/DC-inspired criteria~\cite{SDDFGK18} or other novel criteria such as surprise adequacy \cite{Kim2019aa}. Previous works \cite{pei2017deepxplore,MJX18,SDDFGK18,Kim2019aa}, and future studies on coverage criteria for DNNs could be useful for exposing the defects in DNNs, finding adversarial examples, or forming diverse test sets. 
On the other hand, satisfying a coverage criterion or at least achieving a high coverage measurement can be difficult without a structured method. Existing works \cite{XMJCXLLZYS18,OG18} leverage coverage guided fuzzing (CGF) to achieve high coverage. However, both of these works fuzz inputs randomly. Therefore, their effectiveness is limited, as shown in our experiments. 

In this work, we introduce \methodname, a novel coverage guided fuzzer, for achieving high coverage in DNNs for all existing criteria in the literature. Ultimately, our goal is to help practitioners to extend their test sets with new inputs so that more cases are covered. We leverage Monte Carlo Tree Search (MCTS) \cite{Bouzy2004,chaslot2008progressive}, a search algorithm for decision processes, in achieving this goal. In our method, given an input, MCTS aims to pick a series of mutations that would result in the best coverage increase.

Contributions of this work are as follows:
\begin{itemize}
    \item We introduce a novel advanced coverage guided fuzzing technique for testing DNNs, that is designed to work with every coverage metric.
    \item We show the effectiveness of our method on the most popular coverage criteria and DNNs with changing complexity that are trained on various well-known datasets.
    \item We compare our method with existing coverage guided fuzzing methods in the literature in terms of various metrics.
    \item We also show that some of the inputs generated by our method are adversarial examples.
\end{itemize}

\section{Related Work}

\subsubsection{Coverage-Guided Fuzzing in Software} Fuzzing is a widely used technique for exposing defects in software. Coverage-guided grey-box fuzzing tools, such as AFL\citeyear{afl} and libFuzzer\citeyear{libfuzzer}, have been quite successful in detecting thousands of bugs in many software systems. 

\subsubsection{Testing Deep Neural Networks} Several DNN testing techniques have been developed in the literature recently. Among those, there exist works targeting coverage criteria for DNNs. For example, DeepXplore \cite{pei2017deepxplore} proposed neuron coverage (analogous to statement coverage in software). DeepGauge \cite{MJX18} proposed a set of fine-grained test coverage criteria. DeepCT \cite{ma2019deepct} introduced a combinatorial coverage metric for DNNs. \citet{sun2018testing} presented coverage criteria inspired by MC/DC criteria in software. \citet{Kim2019aa} proposed surprise adequacy criteria based on the amount of surprise caused by the inputs on the neuron activation values. 

\citet{li2019structural} pointed out the limitation of existing structural coverage criteria for neural networks. 

DLFuzz \cite{GJZCS18} introduced the first differential fuzzing framework for deep learning systems. \citet{SWRHKK18} proposed the first concolic testing approach for DNNs. DeepCheck \cite{deepc} tests neural networks based on symbolic execution.  DeepTest \cite{TPSR18} and DeepRoad \cite{ZZZLK18} proposed testing tools for autonomous driving systems. For more studies on testing neural networks, we refer to the work of \citet{zhang2019machine} that surveys testing of machine learning systems.

We now discuss the studies that are close to ours. TensorFuzz \cite{OG18} proposed the first CGF for neural networks that aims to increase a novel coverage metric. DeepHunter \cite{XMJCXLLZYS18} is another work exploring CGF for DNNs by leveraging techniques from software fuzzing, such as power scheduling. 
Our work is distinguished from TensorFuzz and DeepHunter since we use Monte Carlo Tree Search (MCTS) for fuzzing while they do random fuzzing. Moreover, we apply mutations to small regions of a given input, while they apply the mutation to the whole input. \citet{wicker2018feature} also use MCTS for testing DNNs. Our work has several differences from the work of \cite{wicker2018feature}. First, their objective is to find the nearest adversarial example, while our objective is to increase a given coverage criterion. Second, while they use feature extraction such as SIFT (Scale Invariant Feature Transform) \cite{lowe2004distinctive} to ease the selection of pixels to be mutated, we do not need feature extraction because of our formulation of the problem. Third, our method applies image transformations such as brightness change, contrast change, and blur to the images while their approach is based on pixel-level mutations.

\subsubsection{Adversarial Deep Learning}
\citet{szegedy2013intriguing} first discovered the vulnerability of DNNs to adversarial attacks. Since then, multiple white-box and black-box adversarial attacks have been proposed. The most popular white-box attacks include FGSM \cite{goodfellow2015explaining}, BIM \cite{kurakin2016adversarial} , DeepFool \cite{moosavi2016deepfool}, JSMA \cite{papernot2016limitations}, PGD \cite{madry2017towards} and C\&W \cite{carlini2017towards}. 

Attacking to a DNN in a black-box manner is harder than white-box. There exist black-box attacks based on transferability \cite{papernot2016transferability,papernot2017practical}, gradient estimation \cite{chen2018learning,bhagoji2018practical} and local-search \cite{narodytska2017simple} in the literature.

\section{Background}

\subsection{Coverage Criterion}
A coverage criterion is used to measure what percent of the potential behaviors could be tested by a given set of inputs. For this purpose, a coverage criterion divides the input space for a given system into equivalence classes and calculates how many of the equivalence classes have at least one instance input in a given set of inputs. When an equivalence class has at least one instance input in a given set of inputs, the equivalence class is said to be covered. Thus, a coverage criterion calculates how many of its equivalence classes are covered by a set of inputs. If a coverage criterion properly divides the input space for a given system, covering all equivalence classes corresponds to having at least one test sample triggering different behaviors so that all behaviors could be tested.

\subsection{Coverage Guided Fuzzing}
A Coverage Guided Fuzzer (CGF) performs systematic mutations on inputs and produces new test inputs to increase the number of covered cases for a target coverage metric. A typical CGF process starts with selecting a seed input from the seed pool, then continues with mutating the selected seed a certain number of times. After that, the program under test is run with the mutated seed. If a mutated seed creates an increase in the number of covered cases, CGF keeps the mutated seed in the seed pool. In the meantime,  CGF can also keep track of details of execution, such as execution paths and crash reports so that they could be reported if need. 

\subsection{Monte Carlo Tree Search (MCTS)}
Monte Carlo Tree Search \cite{Bouzy2004,chaslot2008progressive} is a search algorithm for decision processes such as game play. It uses game trees to represent a game. Each node of the game tree represents a particular state in the game. On taking an action, one makes a transition from a node to one of its children. MCTS algorithm is used to pick the most promising action on an arbitrary state of a game. Ultimately, the objective is to find the best path (i.e. best sequence of actions) to follow to win the game.

MCTS process can be broken down into the following four steps: \textbf{Selection:} Starting from the root node $R$, successively select child nodes according to their potentials until a leaf node $L$ is reached. The potential of each child node is calculated by using UCT (Upper Confidence Bound applied to Trees) \cite{Kocsis:2006:BBM:2091602.2091633,chaslot2008progressive}. UCT is defined as $v + e \times \sqrt{\frac{\ln{N}}{n}}$ where $v$ refers to the value of the node, $n$ is the visit count of the node, and N is the visit count for the parent of the node. $e$ is a hyperparameter determining exploration-exploitation trade-off. \textbf{Expansion:} Unless $L$ is a terminal node (i.e. win/loss/draw), create at least one child node $C$ (i.e. any valid moves from node $L$) and pick one of them. \textbf{Simulation:} play the game from node $C$ by picking moves randomly until reaching a terminal condition. \textbf{Backpropagation:} propagates back the result of the play to update values associated with the nodes on the path from $C$ to $R$. The path containing the nodes with the highest values in each layer would be the optimal strategy in the game. 

For practical reasons, after applying the MCTS process on root $R$ for a while, it is useful to pick a new root $R'$ from the child nodes of the current root $R$ and continue the MCTS process on the new root $R'$ such that the MCTS gets one level below and continues searching on a subtree.

\section{Method: \methodname}\label{method}

\subsection{Main Idea}
The core idea of our approach is to make use of reward-guided exploration and exploitation, in other words reinforcement learning, to extend the test set by mutating inputs in a smarter way so that to achieve a higher coverage score. In reward-guided process, selected mutations are evaluated with the coverage changes they induce. In this way, our approach is able follow the steps that result in coverage increase. 

\subsection{Notations}
Let $T = \{[X_1, X_2,...], [Y_1, Y_2,...]\}$ be a test set where $(X_i, Y_i)$ is an input-output pair of the $i^{th}$ test sample. Let $\mathcal{I}$ be a set of inputs called a batch. Let $\mathcal{I'}$, $\mathcal{I''}$, ..., $\mathcal{I^{\textit{(n)}}}$ be a sequence of mutated batches of the original batch $\mathcal{I}$ such that:
\begin{equation}
    \mathcal{I} \xrightarrow{\mathcal{IM}(r', m')} \mathcal{I'} \xrightarrow{\mathcal{IM}(r'', m'')} \mathcal{I''}  ... \xrightarrow{\mathcal{IM}(r^{\textit{(n)}}, m^{\textit{(n)}})} \mathcal{I^{\textit{(n)}}}
\end{equation}
where $\mathcal{IM}(r^{\textit{(n)}}, m^{\textit{(n)}})$ is the $n^{th}$ input mutator, $r^{\textit{(n)}}$ and $m^{\textit{(n)}}$ are the region and mutation indexes, respectively. Furthermore, let $\mathcal{I^\textit{best}} \in \{\mathcal{I'}, \mathcal{I''},...\}$ be the best batch, which is the one that creates the greatest amount of coverage increase.

\begin{figure}[hbt!]
    \centering
    \includegraphics[width=\linewidth]{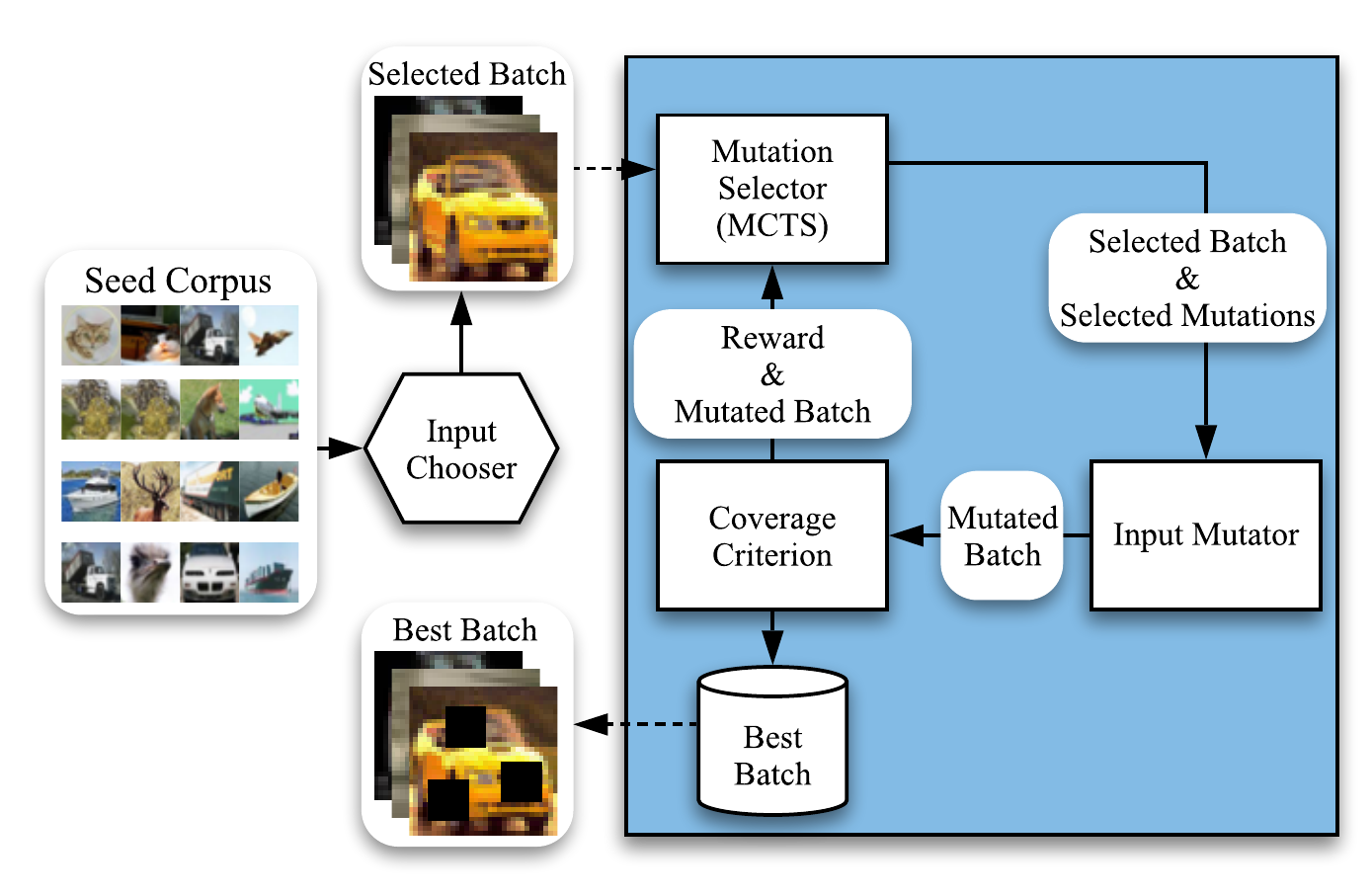}
    \caption{Workflow of {\methodname} for an iteration}
    \label{fig:overview}
\end{figure}

\subsection{Overview}
{\methodname} is a MCTS-driven coverage-guided fuzzer for DNNs. It can be classified as a grey-box testing method since it uses coverage information which is related to internal states of a DNN model.
The method is designed to generate inputs that increase the current level of coverage formed by the initial test set. It is composed of input chooser, coverage criterion, input mutator, and mutation selector, which is the most important part. The coverage criterion is a hyperparameter for our method. For each iteration, the input chooser chooses a batch, which is a set of inputs $\mathcal{I}$. After that, the mutation selector determines which mutation $(r', m')$ to be applied to the inputs. The batch $\mathcal{I}$ and selected mutation $(r', m')$ then goes to input mutator which applies the selected mutation to the batch of inputs so that the mutated inputs $\mathcal{I'}$ are formed ($\mathcal{I} \xrightarrow{\mathcal{IM}(r', m')} \mathcal{I'}$). The mutated inputs are then given to a coverage criterion to calculate the coverage of the mutated inputs union the test set. The coverage and mutated inputs are given to the mutation selector such that it could make use of the coverage and continue working with $\mathcal{I'}$ so that new mutated inputs $\mathcal{I''}$ are generated ($\mathcal{I'} \xrightarrow{\mathcal{IM}(r'', m'')} \mathcal{I''}$). This process continues until a termination condition such as exploration limit or mutation limit is reached. The best set of mutated inputs $\mathcal{I^\textit{best}}$ is stored and updated in the meantime. If there is an increase in coverage because of the mutated inputs $\mathcal{I^\textit{best}}$, they are added to the test set. This concludes the iteration for the batch $\mathcal{I}$. We continue iterating with different batches until a termination condition such as a target number of new inputs or timeout is reached. The workflow of the proposed method is illustrated in Figure \ref{fig:overview}.

\subsection{Input Chooser}
We use two types of input chooser for selecting inputs to make mutations on. These are random input chooser and clustered random input chooser. The random input chooser randomly samples a batch of inputs from the uniform random distribution. The clustered random input chooser aims to sample similar inputs together. It applies an off-the-shelf clustering algorithm. After clustering, it selects a random cluster using the uniform random distribution. Finally, it samples a random batch of inputs from the selected cluster. We use sampling without replacement to avoid multiple same inputs in a batch since we apply the same mutations to all inputs in the batch. In this work, we use k-means clustering as the clustering algorithm.

\subsection{Mutation Selector}
The mutation selector takes a batch of inputs and sequentially selects parameters region index $r$ and mutation index $m$. The selected mutations are sequentially applied to the selected regions by the input mutator and a sequence of mutated batches $\mathcal{I'},\mathcal{I''},..., \mathcal{I^{\textit{(n-1)}}}, \mathcal{I^{\textit{(n)}}}$ is generated. Note that $\mathcal{I^{\textit{(n)}}}$ contains all the mutations up to that point.

\subsubsection{Modelling the mutation selection as a two-player game}
Our proposed mutation selector is a two-player game such that Player I selects the region to be mutated, and Player II selects the mutation to be made on the chosen region. Since regions and mutations are enumerated, these are just integer selection problems. The game continues as Players I and II play iteratively so that multiple local mutations could be applied. 

Region selection and mutation selection are considered as separate actions. We call a tuple of actions taken by Players I and II together as a complete action $(r, m)$.

\subsubsection{Reward}

A naive reward for our problem is the coverage increase for each action.  We use this reward to guide the search for mutations.
In this study, coverage increase corresponds to the difference between the coverage led by current state of the test set and the coverage obtained by adding a new batch to the test set. The purpose of the mutation selector is to find the best mutations that could result in the greatest amount of coverage increase. 

\subsubsection{End of the game}
In order to avoid creating unrealistic inputs, we put some constraints to limit the mutations. 
Generally, $L_p$ norms are used for this purpose. These are defined as $||X_i' - X_i||_p < \epsilon $ where $X_i'$ is the mutated input such that the distance between mutated inputs and original inputs are limited. In general form, let $d(X_i', X_i)$ be a distance metric, the game is over when $d(X_i', X_i) < \epsilon$ constraint is violated, where $d$ and $\epsilon$ are hyperparameters. 

\subsubsection{Searching}
\begin{figure*}[hbt!]
    \centering
    \includegraphics[width=0.8\textwidth, trim=0cm 0.75cm 0cm 0cm]{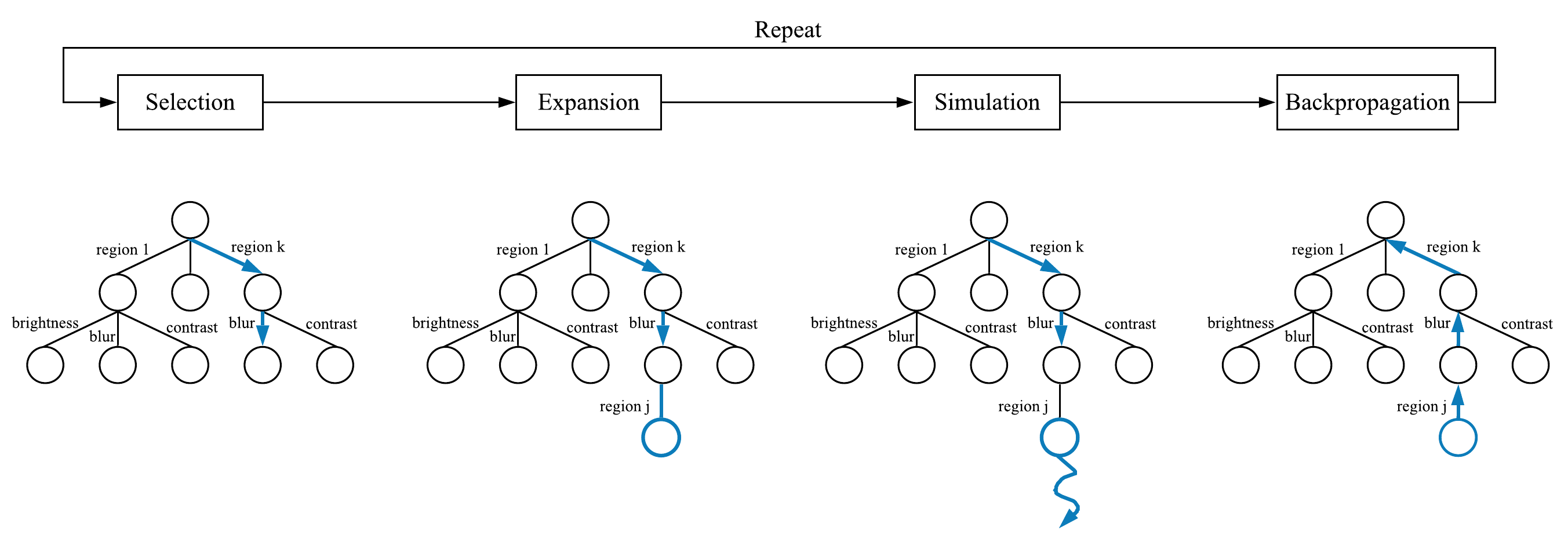}
    \caption{Scheme of a Monte-Carlo Tree Search.}
    \label{fig:mcts}
\end{figure*}

We use Monte Carlo Tree Search (MCTS) in order to search the game tree for the best mutations. The nodes in our game tree correspond to region and mutation selections. We continuously update the batch of inputs $\mathcal{I^\textit{best}}$ that creates the best coverage increase so that the batch is added to the test set when our MCTS is finished. Figure \ref{fig:mcts} shows the steps of our MCTS where edges on odd levels of the game tree correspond to region selections, and edges on even levels of the game tree correspond to mutation selections.

\subsection{Input Mutator}
\begin{figure}[hbt!]
    \centering
    \includegraphics[scale=0.5]{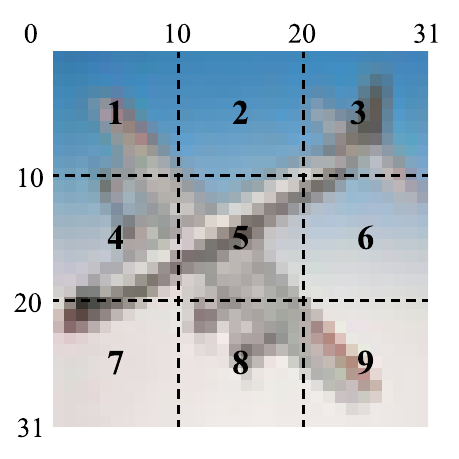}
    \caption{Regions for an example input}
    \label{fig:division}
\end{figure}

The input mutator mutates the input according to the region index and mutation index selected by the mutation selector. The availability of so many mutations could potentially harden the job of Mutation Selector. Therefore, by addressing that problem, we come up with a general input mutator for images. It divides an image into local regions and provides general image mutations as mutation options for each region. These general mutations include, but are not limited to, brightness change, contrast change, and blur. When a region $r$ and a mutation $m$ are selected, it applies the selected mutation to the selected region. The number of regions and the set of available mutations for regions are hyperparameters for the input mutator. With appropriate settings, we can obtain either a pixel-level mutator or an image-level mutator or something in-between, which we think is the best for practical reasons. We enumerate the regions and mutations so that the mutation selector identifies them by indexes. Figure \ref{fig:division} shows an example for the division of input into regions. Our proposed input mutator induces a bias towards locality since it applies mutations to regions of an image. Therefore, it is a natural fit for image problems and convolutional neural networks. 

\subsection{Algorithm}

\begin{algorithm*}[tbh]
\caption{Algorithmic description of our method: \methodname}
\label{alg:method}
\begin{algorithmic}[1]
\Procedure{\methodname}{T, Input\_Chooser, Coverage\_Criterion, $tc_0$, $tc_1$, $tc_2$, $tc_3$}
\While{not $tc_0$}
    \State $\mathcal{I}$ = Input\_Chooser(T)
    \State $R$ = MCTS\_Node($\mathcal{I}$)
    \State best\_coverage, $\mathcal{I^\textit{best}}$ = 0, $\mathcal{I}$
    \While{not $tc_1$}
        \While{not $tc_2$}
            \State $L$ = MCTS\_Selection($R$)
            \State $C$ = MCTS\_Expansion($L$)
            \State $\mathcal{I^{\textit{(n-1)}}}$ = $C$.$\mathcal{I}$ \Comment{$\mathcal{I} \xrightarrow{\textit{MCTS Selection and Expansion}} \mathcal{I^{\textit{(n-1)}}}$}
            \State $r^{\textit{(n)}}, m^{\textit{(n)}}$ = MCTS\_Simulation($C$)
            \State $\mathcal{I^{\textit{(n)}}}$ = Input\_Mutator($\mathcal{I^{\textit{(n-1)}}}$, $r^{\textit{(n)}}, m^{\textit{(n)}}$) \Comment{$
            \mathcal{I^{\textit{(n-1)}}}
            \xrightarrow{\mathcal{IM}(r^{\textit{(n)}}, m^{\textit{(n)}})} \mathcal{I^{\textit{(n)}}}$}
            \If{not $tc_3$} \Comment{the game is not finished: $d(\mathcal{I^{\textit{(n)}}}, \mathcal{I}) < \epsilon$}
                \State coverage\_increase = Coverage\_Criterion(T $\cup$ $\mathcal{I^{\textit{(n)}}}$) - Coverage\_Criterion(T)
                \If{coverage\_increase $>$ best\_coverage}
                \State best\_coverage, $\mathcal{I^\textit{best}}$ = coverage\_increase, $\mathcal{I^{\textit{(n)}}}$
                \EndIf
                \State MCTS\_Backpropagation($C$, coverage\_increase)
            \EndIf
        \EndWhile
        \State $R$ = select\_child($R$) \Comment{selects the child that has the greatest value in terms of coverage increase}
    \EndWhile
    \If{best\_coverage $>$ 0}
        \State T = T $\cup$ $\mathcal{I^\textit{best}}$
    \EndIf
\EndWhile
\State \Return T
\EndProcedure
\end{algorithmic}
\end{algorithm*}

We describe our method in Algorithm \ref{alg:method}. The while loop in line 2 refers to iterating until a termination condition ($tc_0$) that is a timeout or reaching a target number of new inputs. In line 3, a batch of inputs $\mathcal{I}$ is sampled using the input chooser. The root node $R$ is created in line 4, and variables to store best mutated batch $\mathcal{I^\textit{best}}$ are initialized in line 5. The while loop in line 6 refers to looping until a termination condition ($tc_1$) that determines at most how many levels in the search tree the MCTS can go down. Next, the while loop in line 7 refers to iterating until a termination condition ($tc_2$) that determines at most how many times a MCTS node can be explored. In line 8, MCTS Selection is made and it results in a leaf node $L$. Then, in line 9, MCTS Expansion is applied and it creates a new child node $C$ as a child of the leaf node $L$. We assume our MCTS Selection and MCTS Expansion functions mutate the original batch $\mathcal{I}$ according to the game tree and store the batch corresponding to each $node$ as a property of the node, which is denoted as $node.\mathcal{I}$. Note that $node.\mathcal{I}$ is created by applying the mutations corresponding to the edges that are on the path from the root node $R$ to the given $node$. In line 10, the mutated batch that is the result of MCTS Selection and Expansion is referred as $\mathcal{I^{\textit{(n-1)}}}$. Then, the MCTS Simulation plays the game until a complete action so that $r^{(n)}$ and $m^{(n)}$ are assigned to a region index and a mutation index, respectively (line 11). The input mutator then mutates the batch $\mathcal{I^{\textit{(n-1)}}}$ according to a region index $r^{(n)}$ and a mutation index $m^{(n)}$ so that a new batch $\mathcal{I^{\textit{(n)}}}$ is created (line 12). Termination condition ($tc_3$) limits distance between the mutated batch $\mathcal{I^{\textit{(n)}}}$ and the original batch $\mathcal{I}$. If this new batch $\mathcal{I^{\textit{(n)}}}$ does not violate the termination condition ($tc_3$), then the mutated batch $\mathcal{I^{\textit{(n)}}}$ is considered as a candidate batch of test inputs (line 13-14). In line 15, coverage increase is calculated from the difference between coverage of test set $T$ union mutated batch $\mathcal{I^{\textit{(n)}}}$ and just test set $T$. If this is the greatest coverage increase for this batch $\mathcal{I}$, the mutated batch $\mathcal{I^{\textit{(n)}}}$ is stored as the best mutated batch $\mathcal{I^\textit{best}}$ (line 15-16). MCTS Backpropagation is applied from the new child node $C$ with coverage increase as reward (line 17). This concludes one iteration of the while loop with $tc_2$, and the algorithm continues looping this way to explore the root node until $tc_2$. When termination condition $tc_2$ is reached, it sets the best child $R'$ of root $R$ as the new root node (line 18). The best child $R'$ is the node with the greatest value, which is average coverage increase (reward) found on the paths (sequences of mutations) that contain the node. After setting a child as the new root, an iteration of the while loop with $tc_1$ is completed, and the while loop continues iterating by working on the subtree of the previous game tree. After the while loop is completely finished, the best batch found $\mathcal{I^\textit{best}}$ is added to the test set if it creates a coverage increase (line 19-20). This concludes a complete iteration of MCTS on the batch $\mathcal{I}$ and the algorithm continue iterating with new batches until termination condition $tc_0$ is reached. When it is reached, the final test set which includes the mutated inputs found up to that point is returned (line 21). 
\begin{figure*}[hbt!]
    \centering
    \begin{subfigure}{.20\textwidth}
      \centering
      \includegraphics[width=\linewidth]{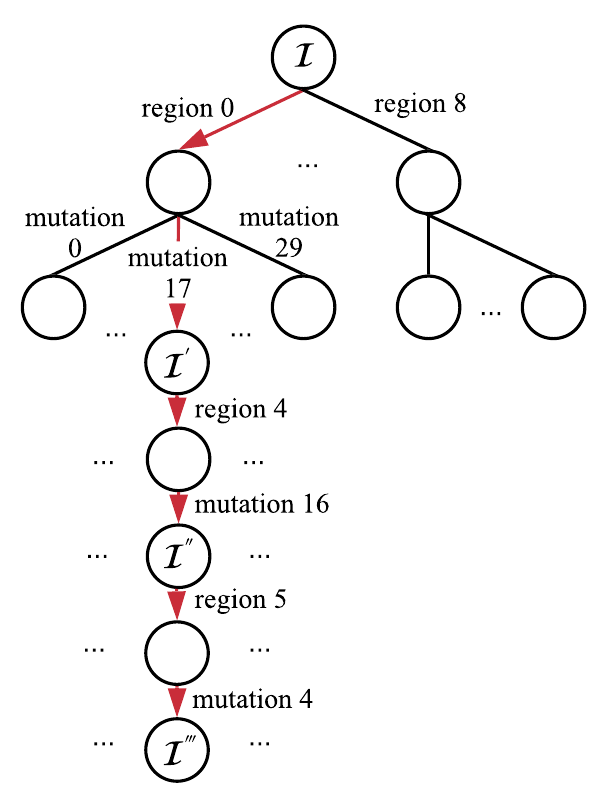}
      \caption{The game tree}
      \label{fig:sub1}
    \end{subfigure}%
    \begin{subfigure}{.65\textwidth}
      \centering
      \includegraphics[width=\linewidth]{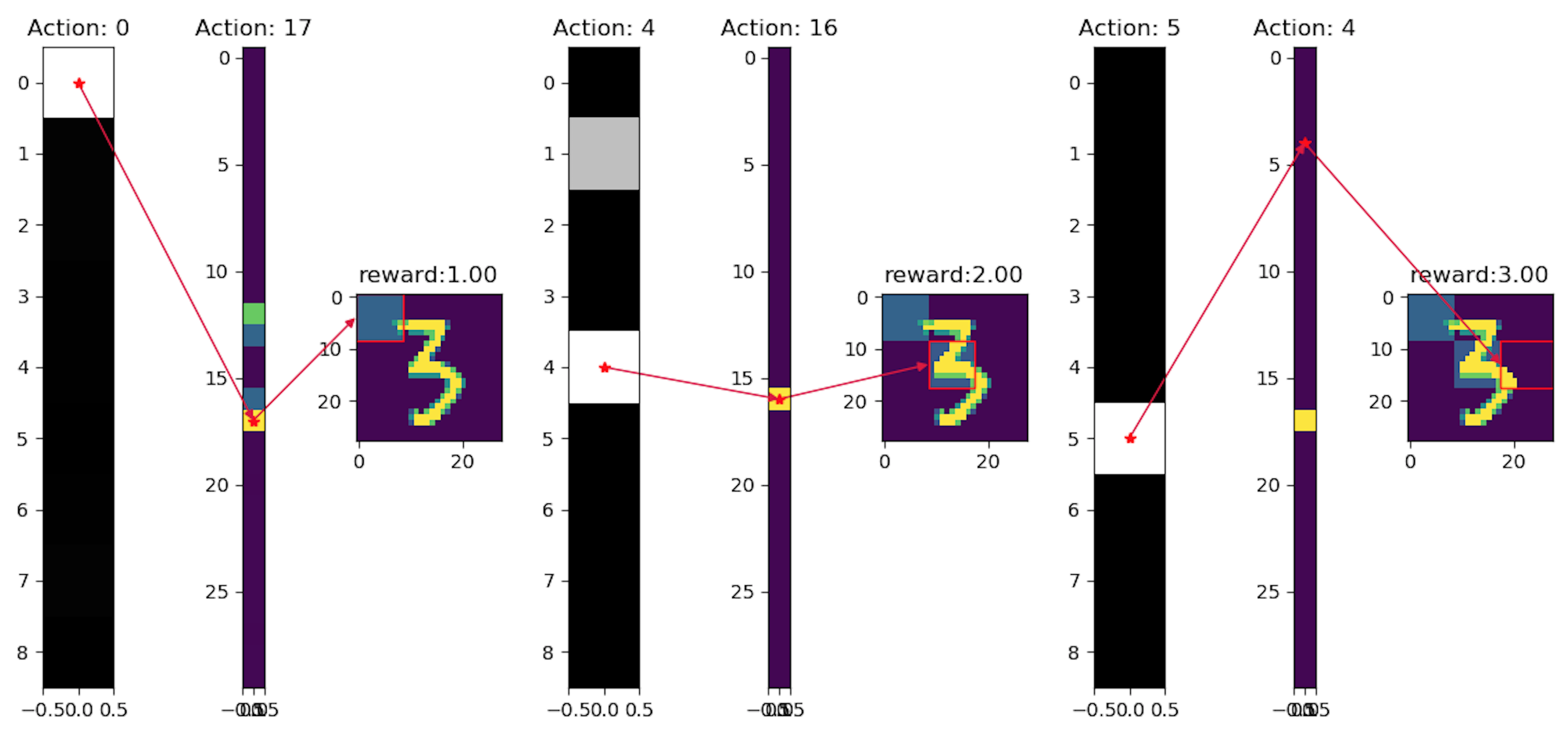}
      \caption{The selected mutations on a seed input}
      \label{fig:sub2}
    \end{subfigure}
    \caption{Visualization of a snapshot when our method searching the mutation space for TFC and MNIST-LeNet5 where action columns represents the potentials (the more brighter the more potential) of each enumerated action on the search tree}
    \label{fig:mcts_in_action}
\end{figure*}
Figure \ref{fig:mcts_in_action} illustrates the algorithm in action as follows: it selects a region (action) and a mutation (action) so that the input mutator applies the mutation to the region, and then this process is continued repeatedly while MCTS is searching the game tree.

\section{Experiments\footnote{The code for our method and experiments can be found on https://github.com/hasanferit/DeepSmartFuzzer.}}
\subsection{Setup}
\subsubsection{Datasets and DL Systems}
We evaluate {\methodname}  on two popular publicly available datasets namely MNIST \cite{lecun1998mnist} and CIFAR10 \cite{krizhevsky2009learning} (referred to as CIFAR from now on). MNIST is a handwritten digit dataset with 60000 training and 10000 testing inputs. Each input is a 28x28 pixel white and black image with a class label from 0 to 9.  CIFAR is a 3-channel colored image dataset with 50000 training and 10000 testing samples. Each input is a 3x32x32 image in ten different classes (e.g., plane, ship, car). For the MNIST dataset, we study LeNet1, LeNet4, and LeNet5 \cite{lecun1998gradient} DNN architectures, which are the three well-known and popular models in the literature. For the CIFAR dataset, we make use of a suggested convolutional neural network (CNN) architecture with proven success to some degree. The subject DNN architectures are selected from different complexities, and they achieve competitive accuracy in the respective test sets. More details can be seen in Table \ref{table:datasets}.

\begin{table}[hbt!]
    \centering
    \caption{Datasets and DL Systems used in our experiments.}
    \label{table:datasets}
    \begin{tabular}{p{1.5cm}p{2.25cm}>{\raggedleft\arraybackslash$}p{1.3cm}<{$}>{\raggedleft\arraybackslash$}p{1.3cm}<{$}}
        \toprule
        \multicolumn{1}{l}{\textbf{Dataset}} & \textbf{DL System} &$\!\!\!$\textbf{Parameters} &$\!\!\!\!\!$\textbf{Accuracy} \\ \midrule
        MNIST & LeNet1 & 7206 & 98.33\% \\ \cmidrule(l){2-4} 
        & LeNet4 & 69362 & 98.59\% \\ \cmidrule(l){2-4} 
        & LeNet5 & 107786 & 98.96\% \\ \midrule
        CIFAR & \begin{tabular}[c]{@{}l@{}}20 layer CNN \\ with max-pooling \\and dropout layers \end{tabular} & 952234 & 77.68\% \\
        \bottomrule
    \end{tabular}
\end{table}

\begin{table*}[hbt!]
\centering
\resizebox{\textwidth}{!}{%
\begin{tabular}{|c|c|c|c|c|c|c|c|c|c|c|}
\hline
\textbf{\begin{tabular}[c]{@{}c@{}}Model - Coverage\\ /\\ CGF\end{tabular}} & \textbf{\begin{tabular}[c]{@{}c@{}}MNIST\\LeNet1\\ (NC)\end{tabular}} & \textbf{\begin{tabular}[c]{@{}c@{}}MNIST\\LeNet1\\ (KMN)\end{tabular}} & \textbf{\begin{tabular}[c]{@{}c@{}}MNIST\\LeNet1\\ (NBC)\end{tabular}} & \textbf{\begin{tabular}[c]{@{}c@{}}MNIST\\LeNet1\\ (SNAC)\end{tabular}} & \textbf{\begin{tabular}[c]{@{}c@{}}MNIST\\LeNet1\\ (TFC)\end{tabular}} & \textbf{\begin{tabular}[c]{@{}c@{}}MNIST\\LeNet4\\ (NC)\end{tabular}} & \textbf{\begin{tabular}[c]{@{}c@{}}MNIST\\LeNet4\\ (KMN)\end{tabular}} & \textbf{\begin{tabular}[c]{@{}c@{}}MNIST\\LeNet4\\ (NBC)\end{tabular}} & \textbf{\begin{tabular}[c]{@{}c@{}}MNIST\\LeNet4\\ (SNAC)\end{tabular}} & \textbf{\begin{tabular}[c]{@{}c@{}}MNIST\\LeNet4 \\ (TFC)\end{tabular}} \\ \hline
{\textbf{DeepHunter}} & 0 & 2.34 $\pm$ 0.03 \% & 35.42 $\pm$ 2.76 \% & 41.67 $\pm$ 4.17 \% & 29.00 $\pm$ 3.61 & 0 & 1.91 $\pm$ 0.04 \% & \textbf{13.15 $\pm$ 2.34 \%} & 16.67 $\pm$ 0.81 \% & 20.00 $\pm$ 2.00 \\ \hline
{\textbf{TensorFuzz}} & 0 & 1.83 $\pm$ 0.23 \% & 0 & 0 & 0.33 $\pm$ 0.58 & 0 & 1.26 $\pm$ 0.05 \% & 0 & 0 & 0 \\ \hline
{\textbf{\methodname}} & 0 & \textbf{2.91 $\pm$ 0.11 \%} & \textbf{41.67 $\pm$ 4.77 \%} & \textbf{42.36 $\pm$ 6.36 \%} & \textbf{204.67 $\pm$ 8.50} & \textbf{1.41 $\pm$ 0.00 \%} & \textbf{2.07 $\pm$ 0.14 \%} & 11.50 $\pm$ 1.13 \% & \textbf{16.90 $\pm$ 3.07 \%} & \textbf{64.33 $\pm$ 6.03} \\ \hline
{\textbf{\methodname (clustered)}} & 0 & 2.88 $\pm$ 0.04 \% & 38.54 $\pm$ 0.00 \% & 39.58 $\pm$ 7.51 \% & 111.00 $\pm$ 14.53 & \textbf{1.41 $\pm$ 0.00 \%} & 2.02 $\pm$ 0.07 \% & 11.50 $\pm$ 0.54 \% & 15.02 $\pm$ 2.15 \% & 53.33 $\pm$ 8.39 \\ \hline
\end{tabular}%
}
\caption{Coverage increase achieved by each CGF for MNIST-LeNet1 and MNIST-LeNet4 models.}
\label{tab:MNIST_LeNet1_4_cov}
\end{table*}

\begin{table*}[hbt!]
\centering
\resizebox{\textwidth}{!}{%
\begin{tabular}{|c|c|c|c|c|c|c|c|c|c|c|}
\hline
\textbf{\begin{tabular}[c]{@{}c@{}}Model - Coverage\\ /\\ CGF\end{tabular}} & \textbf{\begin{tabular}[c]{@{}c@{}}MNIST\\LeNet1\\ (NC)\end{tabular}} & \textbf{\begin{tabular}[c]{@{}c@{}}MNIST\\LeNet1\\ (KMN)\end{tabular}} & \textbf{\begin{tabular}[c]{@{}c@{}}MNIST\\LeNet1\\ (NBC)\end{tabular}} & \textbf{\begin{tabular}[c]{@{}c@{}}MNIST\\LeNet1\\ (SNAC)\end{tabular}} & \textbf{\begin{tabular}[c]{@{}c@{}}MNIST\\LeNet1\\ (TFC)\end{tabular}} & \textbf{\begin{tabular}[c]{@{}c@{}}MNIST\\LeNet4\\ (NC)\end{tabular}} & \textbf{\begin{tabular}[c]{@{}c@{}}MNIST\\LeNet4\\ (KMN)\end{tabular}} & \textbf{\begin{tabular}[c]{@{}c@{}}MNIST\\LeNet4\\ (NBC)\end{tabular}} & \textbf{\begin{tabular}[c]{@{}c@{}}MNIST\\LeNet4\\ (SNAC)\end{tabular}} & \textbf{\begin{tabular}[c]{@{}c@{}}MNIST\\LeNet4 \\ (TFC)\end{tabular}} \\ \hline
{\textbf{DeepHunter}} & 0* & 1051.00 $\pm$ 4.00 & 847.00 $\pm$ 159.74* & 724.67 $\pm$ 180.17* & 1029.67 $\pm$ 29.48 & 0* & 1051.00 $\pm$ 4.00 & 1036.00 $\pm$ 12.49 & 1033.67 $\pm$ 27.50 & 1026.67 $\pm$ 33.50 \\ \hline
{\textbf{TensorFuzz}} & 0* & 1023.33 $\pm$ 1.15 & 0* & 0* & 0.33 $\pm$ 0.58* & 0* & 768.00 $\pm$ 0.00* & 0* & 0* & 0* \\ \hline
{\textbf{\methodname}} & 0* & 1024.00 $\pm$ 0.00 & 1002.67 $\pm$ 36.95 & 533.33 $\pm$ 73.90* & 1024.00 $\pm$ 0.00 & 128.00 $\pm$ 0.00* & 981.33 $\pm$ 73.90$^\dagger$ & 1024.00 $\pm$ 0.00$^\dagger$ & 789.33 $\pm$ 195.52* & 1024.00 $\pm$ 0.00 \\ \hline
{\textbf{\methodname (clustered)}} & 0* & 1024.00 $\pm$ 0.00 & 896.00 $\pm$ 128.00* & 469.33 $\pm$ 97.76* & 1024.00 $\pm$ 0.00 & 128.00 $\pm$ 0.00* & 1024.00 $\pm$ 0.00$^\dagger$ & 1024.00 $\pm$ 0.00$^\dagger$ & 725.33 $\pm$ 97.76* & 1024.00 $\pm$ 0.00 \\ \hline
\end{tabular}%
}
\caption{Number of new inputs produced by each CGF for MNIST-LeNet1 and MNIST-LeNet4 models. *2 hours timeout $^\dagger$Extended experiments with 6 hours limit for timeout}
\label{tab:MNIST_LeNet1_4_inputs}
\end{table*}

\subsubsection{Compared Techniques and Coverage Criteria Benchmarks}
We evaluate our tool by comparing its performance with two existing CGF frameworks for deep learning systems. The first tool, namely DeepHunter \cite{XMJCXLLZYS18}, aims to achieve high coverage by randomly selecting a batch of inputs and applying random mutations on them. DeepHunter also leverages various fuzzing techniques from software testing, such as power scheduling. However, the tool is not publicly available. Therefore we use our implementation of DeepHunter in evaluation. The second tool, namely Tensorfuzz \cite{OG18}, uses the guidance of coverage to debug DNNs. For example, it finds numerical errors and disagreements between neural networks and quantized versions of those networks. Tensorfuzz code is publicly available, and we integrate it into our framework.

For an unbiased evaluation of \methodname, we test our tool on various coverage criteria from the literature. We use DeepXplore's \cite{pei2017deepxplore} neuron coverage (NC), DeepGauge's \cite{MJX18} k-multisection neuron coverage (KMN), neuron boundary coverage (NBC), strong neuron activation coverage (SNAC) and Tensorfuzz's coverage (TFC). Neuron Coverage is defined as the ratio of neurons whose activation value is greater than a threshold to all neurons given a set of inputs. KMN, NBC, and SNAC are derived from Neuron Coverage. KMN divides the space of activation values into k sections. Given a set of inputs, it finds the sections that the activation values of each neuron fall into. Those sections are said to be covered. Then it calculates the ratio of the covered sections to all sections. 
NBC finds the boundaries (maximum and minimum) of activation values of each neuron given the set of training inputs. Neurons whose activation values are out of boundary values are said to cover the boundaries. The ratio of covered boundaries to all boundaries is the coverage value. 
TFC uses the output activation vector of the penultimate layer of the DNN. Different inputs can generate different activation vectors. It differentiates between the activation vectors by using a distance threshold. If a given input creates an activation vector that is distant from the previous inputs, the given input is said to cover a new case, so the coverage is increased by one.  

\subsubsection{Hyperparamters}
We set neuron activation threshold to $0.75$ in NC and the number of sections $k$ to $10000$ in KMN, respectively. For NBC and SNAC, we set as lower (upper) bound the minimum (maximum) activation value encountered in the training set, respectively. These are the recommended settings in the original studies. On the other hand, we observed that the distance threshold used in the original TensorFuzz study was too small for MNIST and CIFAR models such that every little mutation could increase the coverage. Therefore, we tune the threshold of TFC for LeNet1, LeNet4, LeNet5 and CIFAR CNN as $30^2$, $13^2$, $11^2$ and $3^2$, respectively. 

The number of regions, the set of mutations, and termination conditions ($tc_1$, $tc_2$, $tc_3$) constitute the hyperparameters of {\methodname}. The number of regions is selected as 9, which corresponds to $3 \times 3$ division of an image. The set of mutations is selected as contrast change, brightness change, and blur. The first termination conditions ($tc_1$) is chosen to limit MCTS from going down more than 8 levels deep in the game tree. The second termination condition ($tc_2$) limits the number of iterations on each root to 25. For the last termination condition ($tc_3$), we are using the limitations that DeepHunter \cite{XMJCXLLZYS18} puts on the distance between mutated and seed inputs to avoid unrealistic mutated inputs.

\subsection{Results}

\begin{table*}[hbt!]
\centering
\resizebox{\textwidth}{!}{%
\begin{tabular}{|c|c|c|c|c|c|c|c|c|c|c|}
\hline
\textbf{\begin{tabular}[c]{@{}c@{}}Model - Coverage\\ /\\ CGF\end{tabular}} & \textbf{\begin{tabular}[c]{@{}c@{}}MNIST\\ LeNet5\\ (NC)\end{tabular}} & \textbf{\begin{tabular}[c]{@{}c@{}}MNIST\\ LeNet5\\ (KMN)\end{tabular}} & \textbf{\begin{tabular}[c]{@{}c@{}}MNIST\\ LeNet5\\ (NBC)\end{tabular}} & \textbf{\begin{tabular}[c]{@{}c@{}}MNIST\\ LeNet5\\ (SNAC)\end{tabular}} & \textbf{\begin{tabular}[c]{@{}c@{}}MNIST\\ LeNet5\\ (TFC)\end{tabular}} & \textbf{\begin{tabular}[c]{@{}c@{}}CIFAR\\ CNN\\ (NC)\end{tabular}} & \textbf{\begin{tabular}[c]{@{}c@{}}CIFAR\\ CNN\\ (KMN)\end{tabular}} & \textbf{\begin{tabular}[c]{@{}c@{}}CIFAR\\ CNN\\ (NBC)\end{tabular}} & \textbf{\begin{tabular}[c]{@{}c@{}}CIFAR\\ CNN\\ (SNAC)\end{tabular}} & \textbf{\begin{tabular}[c]{@{}c@{}}CIFAR\\ CNN\\ (TFC)\end{tabular}} \\ \hline
\textbf{DeepHunter} & 0.51 $\pm$ 0.58 \% & 1.77 $\pm$ 0.03 \% & 6.23 $\pm$ 0.55 \% & 8.40 $\pm$ 0.66 \% & 19.00 $\pm$ 1.73 & 1.99 $\pm$ 0.19 \% & 0.98 $\pm$ 0.03 \% & 2.39 $\pm$ 0.64 \% & 4.48 $\pm$ 0.90 \% & 16.00 $\pm$ 2.65 \\ \hline
\textbf{Tensorfuzz} & 0.13 $\pm$ 0.22 \% & 0.75 $\pm$ 0.06 \% & 0.13 $\pm$ 0.22 \% & 0 & 1.33 $\pm$ 0.58 & 0.93 $\pm$ 0.15 \% & 0.13 $\pm$ 0.01 \% & 1.54 $\pm$ 0.19 \% & 2.92 $\pm$ 0.34 \% & 0 \\ \hline
\textbf{\methodname} & 2.16 $\pm$ 0.44 \% & \textbf{1.99 $\pm$ 0.01 \%} & 7.82 $\pm$ 1.06 \% & \textbf{9.03 $\pm$ 1.10 \%} & \textbf{76.33 $\pm$ 5.69} & \textbf{3.51 $\pm$ 0.48 \%} & \textbf{1.38 $\pm$ 0.09 \%} & 2.39 $\pm$ 1.23 \% & 4.91 $\pm$ 2.51 & 42.33 $\pm$ 4.51 \\ \hline
\textbf{\methodname (clustered)} & \textbf{2.29 $\pm$ 0.38 \%} & 1.92 $\pm$ 0.08 \% & \textbf{7.89 $\pm$ 0.72 \%} & 8.40 $\pm$ 1.91 \% & 76.00 $\pm$ 8.89 & \textbf{3.51 $\pm$ 0.37 \%} & 1.33 $\pm$ 0.06 \% & \textbf{3.83 $\pm$ 2.66 \%} & \textbf{8.80 $\pm$ 7.11} & \textbf{48.67 $\pm$ 7.02} \\ \hline
\end{tabular}%
}
\caption{Coverage increase achieved by each CGF for MNIST-LeNet5 and CIFAR-CNN models.}
\label{tab:lenet5_cifar_cov_inc}
\end{table*}

\begin{table*}[hbt!]
\centering
\resizebox{\textwidth}{!}{%
\begin{tabular}{|c|c|c|c|c|c|c|c|c|c|c|}
\hline
\textbf{\begin{tabular}[c]{@{}c@{}}Model - Coverage\\ /\\ CGF\end{tabular}} & \textbf{\begin{tabular}[c]{@{}c@{}}MNIST\\ LeNet5\\ (NC)\end{tabular}} & \textbf{\begin{tabular}[c]{@{}c@{}}MNIST\\ LeNet5\\ (KMN)\end{tabular}} & \textbf{\begin{tabular}[c]{@{}c@{}}MNIST\\ LeNet5\\ (NBC)\end{tabular}} & \textbf{\begin{tabular}[c]{@{}c@{}}MNIST\\ LeNet5\\ (SNAC)\end{tabular}} & \textbf{\begin{tabular}[c]{@{}c@{}}MNIST\\ LeNet5\\ (TFC)\end{tabular}} & \textbf{\begin{tabular}[c]{@{}c@{}}CIFAR\\ CNN\\ (NC)\end{tabular}} & \textbf{\begin{tabular}[c]{@{}c@{}}CIFAR\\ CNN\\ (KMN)\end{tabular}} & \textbf{\begin{tabular}[c]{@{}c@{}}CIFAR\\ CNN\\ (NBC)\end{tabular}} & \textbf{\begin{tabular}[c]{@{}c@{}}CIFAR\\ CNN\\ (SNAC)\end{tabular}} & \textbf{\begin{tabular}[c]{@{}c@{}}CIFAR\\ CNN\\ (TFC)\end{tabular}} \\ \hline
\textbf{DeepHunter} & 86.33 $\pm$ 96.81* & 1051.00 $\pm$ 4.00 & 1021.00 $\pm$ 10.54 & 1021.67 $\pm$ 19.66 & 1034.00 $\pm$ 17.52 & 1047.00 $\pm$ 6.24 & 1035.67 $\pm$ 13.58 & 1031.67 $\pm$ 12.34 & 1049.00 $\pm$ 10.54 & 1042.67 $\pm$ 5.51 \\ \hline
\textbf{Tensorfuzz} & 0.33 $\pm$ 0.58* & 448.00 $\pm$ 0.00* & 0.67 $\pm$ 1.15* & 0* & 1.33 $\pm$ 0.58* & 7.33 $\pm$ 1.15* & 192.00 $\pm$ 0.00 & 21.00 $\pm$ 2.65 & 20.67 $\pm$ 3.21 & 0* \\ \hline
\textbf{\methodname} & 362.67 $\pm$ 73.90* & 1024.00 $\pm$ 0.00$^\dagger$ & 1024.00 $\pm$ 0.00$^\dagger$ & 725.33 $\pm$ 36.95* & 1024.00 $\pm$ 0.00 & 1024.00 $\pm$ 0.00 & 1024.00 $\pm$ 0.00$^\dagger$ & 320.00 $\pm$ 0.00 & 341.33 $\pm$ 36.95 & 1024.00 $\pm$ 0.00 \\ \hline
\textbf{\methodname (clustered)} & 362.67 $\pm$ 73.90* & 1024.00 $\pm$ 0.00 $^\dagger$ & 1024.00 $\pm$ 0.00$^\dagger$ & 682.67 $\pm$ 36.95* & 1024.00 $\pm$ 0.00 & 1024.00 $\pm$ 0.00 & 1024.00 $\pm$ 0.00$^\dagger$ & 341.33 $\pm$ 36.95 & 341.33 $\pm$ 36.95 & 1024.00 $\pm$ 0.00 \\ \hline
\end{tabular}%
}
\caption{Number of new inputs produced by each CGF for MNIST-LeNet5 and CIFAR-CNN models. *2 hours timeout $^\dagger$Extended experiments with 6, 12, 24 hours limits for timeout}
\label{tab:lenet5_cifar_inputs}
\end{table*}

\subsubsection{Summary}
\begin{figure}[hbt!]
    \centering
    \begin{subfigure}{.20\linewidth}
      \centering
      \includegraphics[width=0.65\linewidth]{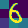}
    \end{subfigure}%
    \begin{subfigure}{.20\linewidth}
      \centering
      \includegraphics[width=0.65\linewidth]{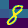}
    \end{subfigure}
    \begin{subfigure}{.20\linewidth}
      \centering
      \includegraphics[width=0.65\linewidth]{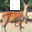}
    \end{subfigure}%
    \begin{subfigure}{.20\linewidth}
      \centering
      \includegraphics[width=0.65\linewidth]{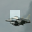}
    \end{subfigure}
    \caption{Example inputs generated by {\methodname}}
    \label{fig:generated_inputs}
\end{figure}

We aim to show that {\methodname} is able to generate good test inputs.  First, we compare {\methodname} with DeepHunter and TensorFuzz by comparing the coverage increases created by approximately 1000 new inputs for each method in combination with different DNN models and coverage criteria. Experimental results show that the inputs generated by our method result in the greatest amount of coverage increase for all (DNN model, coverage criterion) pairs except for a few. This suggests that {\methodname} creates better test inputs than DeepHunter and TensorFuzz with regards to the coverage measurements. 
Second, we calculate the percentage of adversarial, in order words error inducing, test inputs created by our method. The results indicate that there are many adversarial inputs, and they constitute an important part of all generated inputs. Therefore, we conclude that {\methodname} is better than the other coverage-guided fuzzers for DNNs, and it is able to generate adversarial inputs. Figure \ref{fig:generated_inputs} shows two mutated MNIST and two mutated CIFAR inputs created by {\methodname}.

\subsubsection{Comparison to DeepHunter and Tensorfuzz} 
We focus on the inputs generated by {\methodname}, DeepHunter, and Tensorfuzz. For experimental integrity, we make each method generate approximately 1000 input samples. Only the inputs which induce coverage increase are taken into account. We also put a time limit in order to avoid unending cases resulting from being unable to find any coverage increase for some (DNN model, coverage criteria) pairs. When a method could not produce the target amount of inputs in time, yet it creates some coverage increase such that it shows more potential to be explored, the timeout limit is extended so that they could reach 1000 inputs. This condition is not applied to TensorFuzz since it generates inputs one by one, and therefore, it could practically take days to reach 1000 inputs for some cases. The timeout is set to be 2 hours initially. It is then gradually increased to 6, 12, and 24 hours to explore the full potential. Tables \ref{tab:MNIST_LeNet1_4_cov} and \ref{tab:lenet5_cifar_cov_inc} show the amounts of coverage increase produced by approximately 1000 generated input samples from each method with divergent set of coverage criteria and DNN models for MNIST and CIFAR datasets. In order to provide complete results, Tables \ref{tab:MNIST_LeNet1_4_inputs} and \ref{tab:lenet5_cifar_inputs} indicate exactly how many inputs are generated for each case. All of these results are given as mean and standard deviation of the population resulting from running the same experiment three times with different random seeds.
 
For most of the cases, {\methodname} is better than the other two. Especially for the case of TFC, {\methodname} provides a substantial improvement over DeepHunter and TensorFuzz. This might be related to TFC being a layer-level coverage criterion, while the others are neuron-level coverage criteria. Our solution gets better when model complexity is increased. This is suggested by the increasing performance gap between our method and the others. Furthermore, {\methodname} with clustering tends to be better than naive {\methodname} when the complexity of the model is increased. 

On the other hand, for a few cases, our approach fails to provide an improvement. For example, in neuron coverage (NC) with LeNet1 model case, we observe that all fuzzers fail to generate any coverage-increasing input. 
This is because when we cannot find any reward (i.e. coverage increase), our MCTS solution is similar to a random search. 
However, we believe this problem can be avoided with a well-designed reward shaping, and this is left to future work. 
Also, for the case of LeNet4 in combination with NBC, DeepHunter seems to be better than ours. This may indicate a need for further hyperparameter tuning since it conflicts with the general trend. \textbf{Overall, we conclude that {\methodname} provides a significant improvement over existing coverage-guided fuzzers for DNNs.}

\begin{table}[hbt!]
\centering
\resizebox{\linewidth}{!}{%
\begin{tabular}{|c|c|c|c|c|}
\hline
\textbf{\begin{tabular}[c]{@{}c@{}}Model - Coverage\\ /\\Generated Inputs \end{tabular}} & \textbf{\begin{tabular}[c]{@{}c@{}}MNIST\\ LeNet1\\ (KMN)\end{tabular}} & \textbf{\begin{tabular}[c]{@{}c@{}}MNIST\\ LeNet1\\ (NBC)\end{tabular}} & \textbf{\begin{tabular}[c]{@{}c@{}}MNIST\\ LeNet1\\ (SNAC)\end{tabular}} & \textbf{\begin{tabular}[c]{@{}c@{}}MNIST\\ LeNet1\\ (TFC)\end{tabular}} \\ \hline
\textbf{\# Adversarial} & 160.00 $\pm$ 22.11 & 59.00 $\pm$ 3.00 & 32.00 $\pm$ 8.54 & 183.67 $\pm$ 21.78 \\ \hline
\textbf{\# Total} & 917.33 $\pm$ 73.90 & 1002.67 $\pm$ 36.95 & 533.33 $\pm$ 73.90 & 1024.00 $\pm$ 0.00 \\ \hline
\textbf{Percent \%} & \textbf{17.43 $\pm$ 1.73 \%} & \textbf{5.89 $\pm$ 0.37 \%} & \textbf{5.94 $\pm$ 0.92 \%} & \textbf{17.94 $\pm$ 2.13 \%} \\ \hline
\end{tabular}%
}
\caption{Statistics on adversarial inputs generated by {\methodname} on MNIST dataset.}
\label{tab:adversarial}
\end{table}
\vspace{-1em}

\subsubsection{Adversarial Input Generation}
We further experiment with our proposed method in order to check how good it is at finding error-inducing inputs, which is one of the ultimate purposes of testing. The results are provided in Table \ref{tab:adversarial}. The percentage of adversarial inputs that are generated by {\methodname} is dependent on the used coverage criteria as expected. {\methodname} with KMN or TFC provides approximately 17\% adversarial inputs while  {\methodname} with NBC or SNAC provides approximately 5\% adversarial inputs. Although one may think that these percentages are low, we note that adversarial example generation is not the main goal of {\methodname} and the coverage criteria used in this paper.
Thus, we can say that {\methodname} is able to generate error-inducing (adversarial) input samples.
\vspace{-1em}

\section{Conclusion \& Future Work}
In this study, we introduce an advanced coverage guided fuzzer for DNNs that uses Monte Carlo Tree Search (MCTS) to explore and exploit the coverage increase patterns. We experimentally show that our method is better than the previous coverage guided fuzzers for DNNs. Our results also show the potential of reinforcement learning methods for DNN testing. We use naive coverage increase as reward. Therefore, experimentation with reward shaping and different reinforcement learning methods for this problem are left to future studies. We also show that an important portion of the test inputs generated by our method is adversarial. This suggests that our method is successful in terms of testing DNNs. Finally, we share the code for our experiments online in order to provide a base for future studies.

\begin{small}
\bibliographystyle{aaai}
\bibliography{references}
\end{small}

\end{document}